\def\paperID{10108}
\def\confName{CVPR}
\def\confYear{2024}

\def\paperTitle{Leveraging Habitat Information for Fine-grained Bird Identification}

\def\authorBlock{
    Tin Nguyen \textsuperscript{1\footnotemark[1]} \qquad
    Peijie Chen \textsuperscript{1} \qquad
    Anh Nguyen \textsuperscript{1} \qquad \\
    \textsuperscript{1} Auburn University \\
    {\tt\small ttn0011@auburn.edu, \{peijiechen, anh.ng8\}@gmail.com}
}

\newif\ifreview 
\newif\ifarxiv \newcommand{\arxiv}{\arxivtrue}
\newif\ifcamera 
\newif\ifrebuttal

\arxiv 
\pdfoutput=1
\documentclass[10pt,twocolumn,letterpaper]{article}

\ifreview \usepackage[review]{cvpr} \fi
\ifarxiv \usepackage[pagenumbers]{cvpr} \fi
\ifrebuttal \usepackage[rebuttal]{cvpr} \fi
\ifcamera \usepackage{cvpr} \fi

\usepackage{graphicx}	
\usepackage{amsmath}	
\usepackage{amssymb}	
\usepackage{booktabs}
\usepackage{times}
\usepackage{microtype}
\usepackage{epsfig}
\usepackage[table,xcdraw,dvipsnames]{xcolor}
\usepackage{caption}
\usepackage{float}
\usepackage{placeins}
\usepackage{color, colortbl}
\usepackage{stfloats}
\usepackage{enumitem}
\usepackage{tabularx}
\usepackage{xstring}
\usepackage{multirow}
\usepackage{xspace}
\usepackage{url}
\usepackage{subcaption}
\usepackage{xcolor}
\usepackage{multirow}
\usepackage[table]{xcolor}
\usepackage{pifont}
\usepackage{stackengine}
\usepackage{tikz}
\usepackage{makecell}
\usepackage{algorithm}
\usepackage{algpseudocode}
\usepackage{algcompatible}

\usepackage{listings}
\lstdefinestyle{nohighlight}{
}

\usepackage[hang,flushmargin]{footmisc}

\ifcamera \usepackage[accsupp]{axessibility} \fi

% ...

\ifarxiv  \fi

\newcommand{\R}[1]{{%
    \textbf{%
        \ifstrequal{#1}{1}{\textcolor{red}{R#1}}{%
        \ifstrequal{#1}{2}{\textcolor{blue}{R#1}}{%
        \ifstrequal{#1}{3}{\textcolor{magenta}{R#1}}{%
        \ifstrequal{#1}{4}{\textcolor{teal}{R#1}}{%
                           \textcolor{cyan}{R#1}%
        }}}}%
    }%
}}

\newcommand{\increase}[1]{(\textcolor{ForestGreen}{+#1})}
\newcommand{\increasenoparent}[1]{\textcolor{ForestGreen}{+#1}}
\newcommand{\decrease}[1]{(\textcolor{red}{-#1})}

\definecolor{myblue}{rgb}{0.204, 0.596, 0.859}
\definecolor{mygreen}{rgb}{0.180, 0.800, 0.443}
\definecolor{myorange}{rgb}{0.902, 0.494, 0.133}

\newcommand{\AugIrrelevant}{Mixed-Irrelevant}
\newcommand{\AugGroup}{Mixed-Group}
\newcommand{\AugSame}{Mixed-Same}
\newcommand{\AugNone}{Baseline}
\newcommand{\xmark}{\ding{55}}

\newcommand{\allaboutbird}{allaboutbirds.org}
\newcommand{\ssc}{Shape, Size, and Color\xspace}

\newcommand{\cub}{CUB\xspace}
\newcommand{\nabirds}{NABirds\xspace}
\newcommand{\inat}{iNaturalist-birds\xspace}
\newcommand{\inatcub}{iNaturalist-CUB\xspace}
\newcommand{\inatnabirds}{iNaturalist-NABirds\xspace}

\newcommand{\unimodal}{vision-only\xspace}
\newcommand{\multimodal}{vision-language\xspace}

\definecolor{customcolor}{RGB}{100,200,50}

\newcommand{\class}[1]{{\footnotesize\texttt{#1}\xspace}}

\usepackage{xr-hyper}

\makeatletter
\newcommand*{\addFileDependency}[1]{
  \typeout{(#1)}
  \@addtofilelist{#1}
  \IfFileExists{#1}{}{\typeout{No file #1.}}
}

\makeatother

\definecolor{cvprblue}{rgb}{0.21,0.49,0.74}
\usepackage[pagebackref,breaklinks,colorlinks,citecolor=cvprblue]{hyperref}
\usepackage[capitalize]{cleveref}
\crefname{section}{Sec.}{Secs.}
\crefname{table}{Table}{Tables}
\crefname{figure}{Fig.}{Figs.}

\begin{document}
\title{\paperTitle}
\author{\authorBlock}
\maketitle

\begin{abstract}

Traditional bird classifiers mostly rely on the visual characteristics of birds. 
Some prior works even train classifiers to be invariant to the background, completely discarding the living environment of birds.
Instead, we are the first to explore integrating \textbf{habitat} information, one of the four major cues for identifying birds by ornithologists, into modern bird classifiers.
We focus on two leading model types: (1) CNNs and ViTs trained on the downstream bird datasets; and (2) original, multi-modal CLIP \cite{radford2021learning}. 
Training CNNs and ViTs with habitat-augmented data results in an improvement of up to \increasenoparent{0.83} and \increasenoparent{0.23} points on NABirds and CUB-200, respectively. 
Similarly, adding habitat descriptors to the prompts for CLIP yields a substantial accuracy boost of up to \increasenoparent{0.99} and \increasenoparent{1.1} points on NABirds and CUB-200, respectively. 
We find consistent accuracy improvement after integrating habitat features into the image augmentation process and into the textual descriptors of vision-language CLIP classifiers.
Code is available at: \url{https://github.com/tin-xai/habitat}.
\end{abstract}

\section{Introduction}
\label{sec:intro}
\begin{figure}[ht]
  \centering
  \makebox[\columnwidth][c]{\textbf{Test Accuracy on NABirds}}%
    
    \begin{minipage}{0.9\columnwidth}
        \includegraphics[width=\columnwidth]{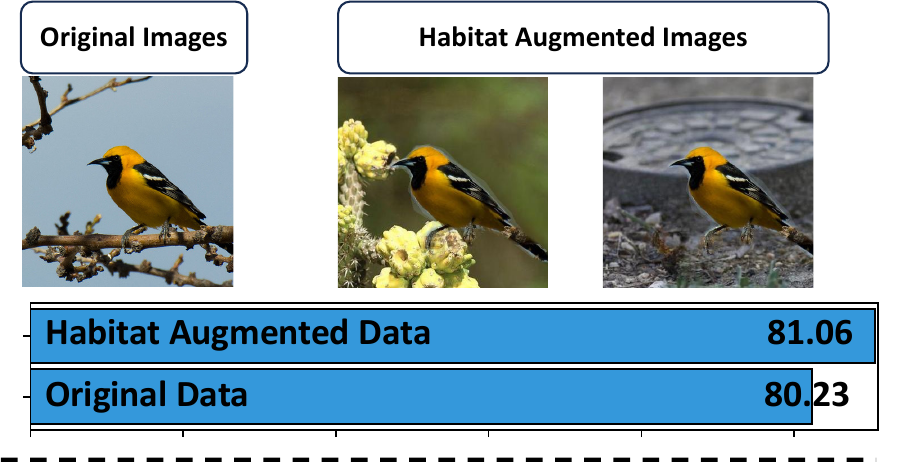}
    \end{minipage} 
    
    \makebox[\columnwidth][c]{\textbf{Zero-shot Accuracy on NABirds}}%
     
    \footnotesize\makebox[0.3\columnwidth][c]{(Ours) Visual \cite{allaboutbirds} + Habitat \cite{allaboutbirds}}%
    \footnotesize\makebox[0.35\columnwidth][c]{Visual \cite{allaboutbirds}}%
    \footnotesize\makebox[0.15\columnwidth][c]{CLIP \cite{radford2021learning}}%
    
    \begin{minipage}{0.9\columnwidth}
        \includegraphics[width=\columnwidth]{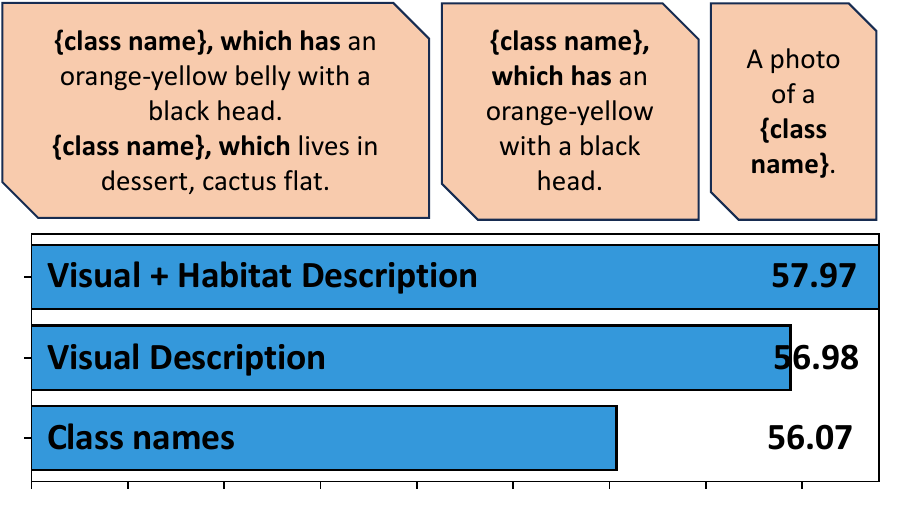}
    \end{minipage}
  
  \caption{\textbf{Top:} CNN trained on with habitat-augmented data improves accuracy by \increasenoparent{0.83} pts over original data. 
  \textbf{Bottom:} Adding habitat descriptions to CLIP boosts zero-shot accuracy, exceeding visually-based or class name-only descriptions by \increasenoparent{0.99} pts and \increasenoparent{1.90} pts, respectively (details in \cref{tab:combined_performance}, \ref{tab:multimodal_cub_nabirds}, and \ref{tab:multimodal_inat}). Note that, both models are tested on \nabirds.}
  \label{fig:overview}
\end{figure}

\begin{figure}[ht]
  \centering
  \begin{subfigure}[t]{0.48\columnwidth}
    \includegraphics[width=\linewidth]{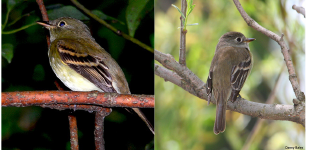}
    \scriptsize
    \begin{tabular}{c@{\hspace{0.2cm}}c}
    \hspace{-0.3cm}\textbf{Acadian Flycatcher} & \textbf{Least Flycatcher} \\
    (a) Swamp & (b) Edges of Woods \\
    \end{tabular}
  \end{subfigure}
  \hfill 
  \begin{subfigure}[t]{0.48\columnwidth}
    \includegraphics[width=\linewidth]{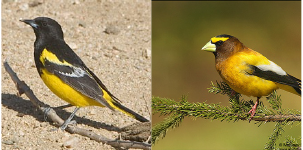}
    \scriptsize
    \begin{tabular}{c@{\hspace{0.4cm}}c}
    \hspace{0.2cm}\textbf{Scott Oriole} & \textbf{Evening Grosbeak} \\
    (c) Desert & (d) Pine-oak Tree \\
    \end{tabular}
  \end{subfigure}

  \caption{Visual comparison of two bird pairs with similar morphologies but different habitats: \class{Acadian Flycatcher} in swamps vs. \class{Least Flycatcher} in woodland edges; \class{Scott Oriole} in deserts vs. \class{Evening Grosbeak} in pine-oak areas. See Appendix \cref{sec:more_bird_pair} for details.}
  \label{fig:overview_2}
\end{figure}

Bird identification, despite recent progress, continues to be challenging, particularly in distinguishing between species with similar appearances, such as various sparrow species \cite{wah2011caltech}. A promising feature that can address the issue is habitat, which is recognized as one of four keys to identifying birds by bird watchers and ornithologists \cite{allaboutbirds_fourkeys,birdsandblooms_habitat,avianreport2023}.
This is exemplified by species like the \class{Scott Oriole} and \class{Evening Grosbeak} (\cref{fig:overview_2}), which are visually similar but inhabit distinct environments: The former nest and forage in deserts or cactus flats, whereas the latter breed in pine-oak, pinyon-juniper, and aspen forests \cite{allaboutbirds}.

Early works focus on finding the most discriminative features of birds, giving rise to part-based models \cite{zheng2017learning,lin2015bilinear,yang2018learning}. More recent efforts have expanded to include additional information like geolocation \cite{chu2019geo,mac2019presence,tang2015improving} and textual descriptions \cite{he2017fine,diao2022metaformer} to enhance identification accuracy. \cite{xiao2020noise} made preliminary efforts to emphasize the significance of background for classification. Additionally, other researchers have approached the problem by focusing on habitat classification \cite{wang2021identifying,garcia2023machine}.

However, to our best knowledge, no prior works explored how to leverage habitat information to improve bird identification accuracy.
Our study addresses this research gap by introducing novel methodologies for integrating habitat information into both \unimodal and \multimodal models, broadening the scope of bird identification techniques. For \unimodal models, we introduce a technique of superimposing bird images onto their relevant habitat backgrounds, creating habitat-augmented data for training purposes. In the \multimodal model context, we specifically tailor the text input to include relevant habitat information, a targeted approach that contributes to improved accuracy in bird recognition.

In this work, we aim to assess how habitat information influences the performance of both \unimodal and \multimodal models in bird identification. We focus on evaluating two prominent \unimodal models, ResNet-50 \cite{he2016deep} trained by \cite{taesiri2022visual}, and TransFG \cite{he2022transfg} with a Vision Transformer (ViT) \cite{dosovitskiy2020image} backbone, each augmented with habitat-specific data. Additionally, for a \multimodal approach, we examine the CLIP model \cite{radford2021learning}, adding habitat details to textual prompts. This methodology aims to highlight the potential of habitat integration in enhancing model accuracy, offering a comprehensive analysis of its impact on ornithological machine learning.

As highlighted by \url{\allaboutbird}, a leading resource in avian studies, critical attributes for identifying bird species include shape \& size, color pattern, behavior, and \textbf{habitat}. Consistent with these attributes, our study demonstrates that the integration of habitat data consistently improves the performance of both \unimodal and \multimodal models in bird identification, thereby validating the significance of this approach. To the best of our knowledge, our work is the first to demonstrate the effectiveness of habitat information in bird identification. 
Our main findings are:

\begin{itemize}
    \item State-of-the-art \unimodal ~models can be improved in accuracy when training with habitat-augmented images (\eg, \increasenoparent{0.23} pts in \nabirds for ResNet-50 as shown in \cref{tab:combined_performance}).
    \item The addition of habitat descriptions to the CLIP model significantly boosts its zero-shot and few-shot accuracy, with increases of up to \increasenoparent{1.1} and \increasenoparent{4.63} points respectively (\cref{tab:multimodal_cub_nabirds,tab:multimodal_inat,tab:few_shot_cub_nabirds}).
    \item State-of-the-art models struggle with images lacking habitat information, such as FlyBird images. In the \nabirds dataset, Non-FlyBird images outperform FlyBird by \increasenoparent{6.92} points in \unimodal and \increasenoparent{4.10} points in \multimodal models, emphasizing the crucial role of habitat data (\cref{tab:uncertain_1,tab:uncertain_2}).
\end{itemize}

\section{Related Work}
\label{sec:related}
Current methods in fine-grained bird classification have two main branches: \unimodal models, which focus on augmenting visual information through innovative architectural designs, and \multimodal models (VLMs), which integrate external textual knowledge as an additional modality. Exploring these branches in detail reveals distinct methodologies and their evolution in bird classification.

\paragraph{Utilizing visual features in \unimodal models}
Early methods in fine-grained bird classification primarily explored visual features, such as zooming into particular regions \cite{fu2017look}, extracting certain visual features \cite{zheng2017learning,lin2015bilinear,yang2018learning}.
recent approaches have advanced this field by concentrating on bird-specific parts or concepts to both enhance and explain the classification process \cite{he2022transfg,nauta2022looks,donnelly2022deformable,Nauta_2021_CVPR,sacha2023protoseg,xue2022protopformer}. These methods have intensively focused on avian characteristics, yet notably, they have not incorporated habitat information into their classification frameworks.

\paragraph{Adding extra information as another modality}
Incorporating additional data types, such as geolocation and time of observation, has proven beneficial in fine-grained bird classification. Researchers have found that geographical information, in particular, can significantly improve classification accuracy \cite{chu2019geo,mac2019presence,diao2022metaformer}. Recently, Visual Language Models (VLMs) like CLIP \cite{radford2021learning}, FILIP \cite{yao2021filip}, and CoCa \cite{yu2022coca}, which combine visual and textual data, have shown promising results. These models, pretrained on large-scale datasets \cite{schuhmann2021laion,schuhmann2022laion}, perform comparably to specialized fine-grained classifiers. \cite{menon2022visual} found that the foundation model CLIP \cite{radford2021learning} can be further amplified by adding more descriptive textual descriptions. However, these methods have yet to fully explore the use of habitat information, a key factor in bird identification \cite{bto2023,marini2019,cornell2023,frontiers2023,sciencedirect2023}.

\paragraph{Background Information}


Several studies have emphasized the influence of background in image classification. For instance, \cite{xiao2020noise} found that changing backgrounds can significantly affect model accuracy, as models exploit background correlations. Similarly, \cite{zhu2016object} demonstrated enhanced object recognition performance by combining networks trained on both the foreground object and the background context. In a different approach, \cite{wang2022clad} developed a contrastive learning method to mitigate background bias. Building on top of these insights, our research takes a distinct path by explicitly emphasizing the recognition of habitat elements in bird identification, an aspect pivotal yet needs to be explored in the field.

\paragraph{Habitat Classification} 

While \cite{wang2021identifying} conducted a study on recognizing habitat elements in bird images, their approach did not directly apply this information to bird species identification. Instead, they developed a classifier for categorizing different types of habitats, such as stone, water, leafless areas, or trunks. Our approach, however, diverges significantly. We integrate habitat information into our dataset through augmentation, enriching the data associated with each bird species. This strategy allows our model to leverage habitat context, making more informed and accurate identification decisions.

\section{Methods}
\label{sec:method}
This section presents our proposed habitat-based data augmentation for \unimodal models in \cref{sec:method_unimodal}. Then, we propose a simple approach to include habitat into bird descriptions for the foundation model CLIP \cref{sec:method_multimodal}.

\begin{figure}[ht]
    \centering
    \scriptsize\makebox[0.2\columnwidth][c]{Original}%
    \scriptsize\makebox[0.65\columnwidth][c]{Augmented}%
    
    \begin{minipage}{0.05\columnwidth}
            \rotatebox{90}{Mixed-S}
            \rule{0pt}{25pt} 
            \rotatebox{90}{Mixed-G}
            \rule{0pt}{40pt} 
            \rotatebox{90}{Mixed-I}
            \rule{0pt}{40pt} 
    \end{minipage}%
    \begin{minipage}{0.9\columnwidth}
    \includegraphics[width=\columnwidth, height=4cm]{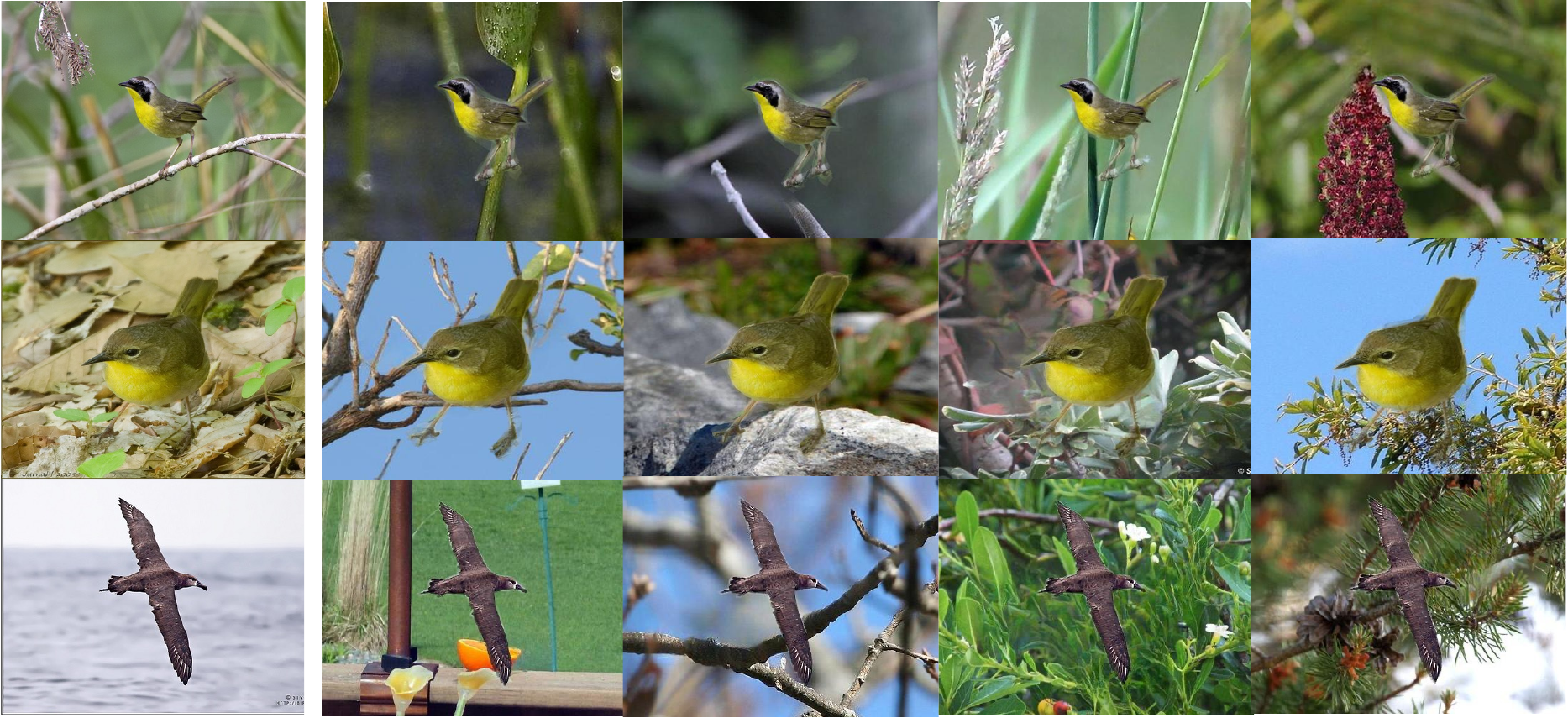}
    \end{minipage}

  \caption{
  Three augmentation techniques (Mixed-S, Mixed-G, Mixed-I) are illustrated: Original bird images in the first column, followed by augmented versions.
  The first row shows \class{Common Yellow Throat} in varying habitats (marsh to grassland). The second row features it amidst different species sharing the same habitats. The last row demonstrates \AugIrrelevant, placing \class{Black Footed Albatross} (typically found near shores) in forest and grass backgrounds.}
  \label{fig:augment_examples}
\end{figure}

\begin{figure}[ht]
  \centering
  \includegraphics[width=\columnwidth]{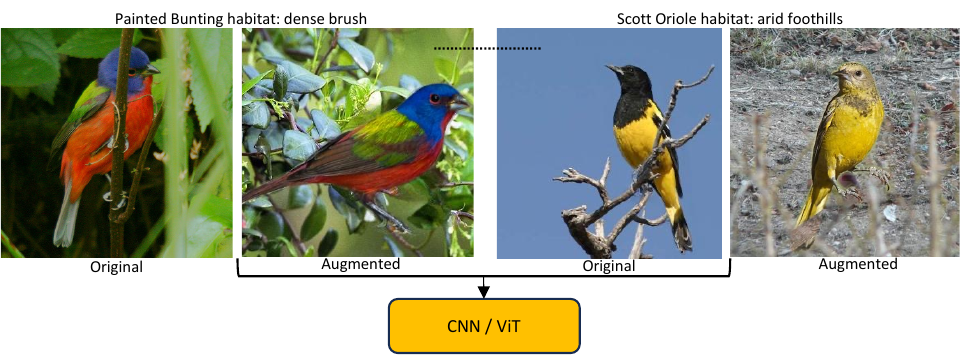}
  
  \caption{Bird classification with \unimodal models (CNN, ViT) utilizes augmented datasets blending original and habitat-augmented images. The augmented images have more contextual habitat, for instance, the habitats of \class{Painted Bunting} and \class{Scott Oriole} are dense brush and arid foothills.}
  \label{fig:unimodal}
\end{figure}

\subsection{Improve Habitat Understanding in Vision-Only Models with Habitat-augmented Data}
\label{sec:method_unimodal}

For \unimodal models, we evaluate the effectiveness of our habitat-based augmentation methods on two representative models (\cref{fig:unimodal}): ResNet-50 \cite{taesiri2022visual} for CNN and TransFG \cite{taesiri2022visual} for ViT, which employs ViT-B/16 \cite{dosovitskiy2020image} as its backbone.

We introduce three augmentation methods: \textbf{\AugSame (Mixed-S)} and \textbf{\AugGroup (Mixed-G)}, inspired by \cite{xiao2020noise,ghiasi2021simple}, aimed to enhance habitat diversity. Moreover, \textbf{\AugIrrelevant (Mixed-I)}, designed to reduce habitat correlation with birds, serves as a control in evaluating the impact of habitat diversity on bird identification.

The essence of our augmentation approach lies in isolating the bird from an image and seamlessly integrating it into a distinct habitat image. To achieve this, we generate two specialized image collections. The first, \textbf{Only Bird Images}, in which birds against a uniform black background; the second, \textbf{Habitat Images}, images of avian environments with birds removed by LaMa \cite{suvorov2022resolution}. We provide an example of these images in \cref{fig:adversarial_examples}. Subsequently, these curated sets allow us to overlay the \textbf{Only Bird Images} onto the \textbf{Habitat Images}.  This superimposition can include habitats that are either congruent with the bird's class, aligned with its broader taxonomic group, or derived from an entirely different grouping.

\paragraph{\AugSame \xspace (Mixed-S)} We randomly change the background of a bird within its class, this method is inspired by previous works \cite{xiao2020noise,ghiasi2021simple}. Specifically, this entails superimposing each ``Only Bird Image" onto a corresponding ``Habitat Image" that shares the same class index. This process ensures that the bird is depicted against a variety of backgrounds that are class-consistent, thereby enriching the dataset with diverse, yet relevant, environmental contexts. An example of this process for the Common Yellow Throat and Black-footed Albatross is given in \cref{fig:augment_examples}.

\paragraph{\AugGroup \xspace (Mixed-G)} In light of the fact that certain avian species share habitats, our approach involves categorizing birds into clusters based on the similarity of their environmental settings, concretely, we use k-means \cite{lloyd1982least} to cluster birds into groups based on their habitat descriptions sourced from \allaboutbird \cite{allaboutbirds} (see algorithm details in Appendix \cref{sec:constructing_additional_data,sec:app_description}). This insight has led to the development of a novel augmentation technique, termed \AugGroup, which strategically superimposes each ``Only Bird Image" onto a ``Habitat Image" within the same ecological group.

For instance, as illustrated in the second row in \cref{fig:augment_examples}, the \class{Common Yellow Throat} is placed onto different habitats of a group including \class{Orange-crowned Warbler}, \class{Savannah Sparrow}, \class{Groove-billed Ani}, and \class{Florida Jay}, these birds share the same habitat preference for deciduous shrub environments.

We provide some details of habitat groups according to habitat descriptions in Appendix \cref{sec:constructing_additional_data}.

\paragraph{\AugIrrelevant \xspace (Mixed-I)}To understand the importance of habitats to bird recognition, a bird will be isolated from its own habitat or habitat group and placed into a completely irrelevant one. 
In \AugIrrelevant, we randomly choose a habitat image from all classes that are not in the habitat groups of a specific bird to do image augmentation. \AugIrrelevant injects irrelevant habitat to the training images, therefore enforcing the model to utilize the bird information only. As can be seen from Fig. \ref{fig:augment_examples}, after augmentation, the \class{Black-footed Albatross} is placed in a forest which is completely unnatural because its habitats are on low, sandy islands, or near shore, and far offshore.


\subsection{Habitat Understanding in Multimodal Models}
\label{sec:method_multimodal}
\begin{figure}[ht]
  \centering
    \includegraphics[width=\columnwidth]{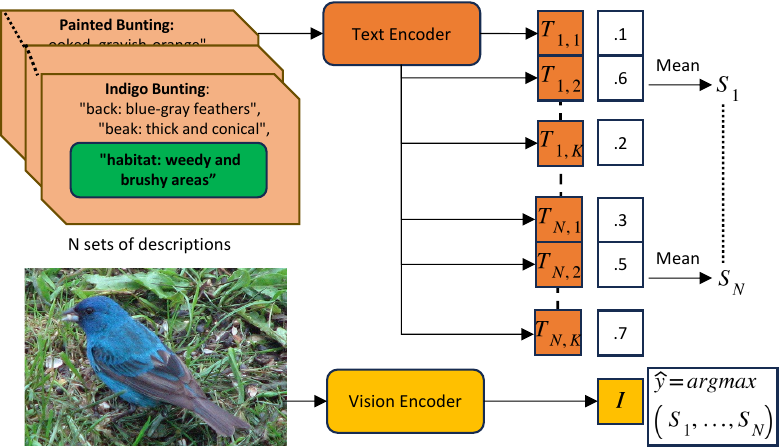}
  
  \caption{Integrating habitat data into CLIP during zero-shot enhances bird identification. Each class comes with descriptions; CLIP calculates and averages similarity scores between these and the input image. The class with the highest softmax score is then predicted.}
  \label{fig:clip}
\end{figure}

In terms of multimodal models, we use CLIP (\cref{fig:clip}) and investigate its zero-shot capability in the presence of supplementary habitat descriptions. 

Previously, M\&V \cite{menon2022visual}, or PEEB \cite{anonymous2023partbased} employed fine-grained descriptions generated by LLMs as input prompts to CLIP. Although these descriptors provide rich information about bird appearance such as color, or shape, they still lack descriptors describing habitat.

Motivated by this, we employ habitat descriptors sourced from \allaboutbird \cite{allaboutbirds}, providing valuable insights into the environments where each bird species thrives. 

\section{Results}
\label{sec:experiment}
\begin{figure}[ht]
  \centering
  \includegraphics[width=\columnwidth]{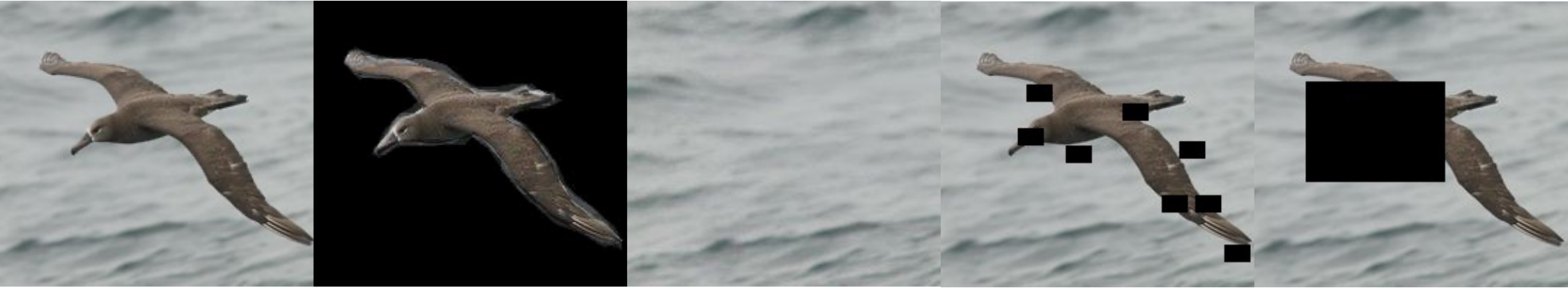}
  \scriptsize\makebox[0.2\columnwidth][c]{Original}%
    \scriptsize\makebox[0.2\columnwidth][c]{Black Background}%
    \scriptsize\makebox[0.2\columnwidth][c]{No Bird}%
    \scriptsize\makebox[0.2\columnwidth][c]{Black Boxes}%
    \scriptsize\makebox[0.2\columnwidth][c]{Big Box}%
  \caption{Five test scenarios assess model robustness with variations: no habitat (black background), habitat only (no bird), and obscured bird images.}
  \label{fig:adversarial_examples}
\end{figure}

To establish the broad applicability and impact of the habitat, we evaluate \unimodal models across diverse test sets (refer to \cref{fig:adversarial_examples}) and assess CLIP's performance with various description sets (detailed in \cref{sec:clip_exp}). All tests were performed using PyTorch \cite{paszke2019pytorch} on a single NVIDIA V100 GPU.

\begin{table*}
\scriptsize
    \centering
    \caption{
    Accuracy comparison across various \cub and \nabirds test sets using ResNet-50 \cite{taesiri2022visual} and TransFG \cite{he2022transfg} models shows \AugSame and \AugGroup's effectiveness, especially in challenging scenarios like \textbf{No Birds, Small Boxes, and Big Box}, indicating superior utilization of habitat information.}
    \begin{tabular}{|c|l|c|c|c|c|c|c|c|c|}
    \hline

     \multicolumn{2}{|c|}{\multirow{2}{*}{\textbf{Test Data}}} & \multicolumn{4}{c|}{\textbf{ResNet-50} \cite{taesiri2022visual}} & \multicolumn{4}{c|}{\textbf{TransFG (ViT/B-16)} \cite{he2022transfg}} \\
    \cline{3-10}
     \multicolumn{2}{|c|}{} & \AugNone & \makecell{Mixed-I} & Mixed-G & Mixed-S & \AugNone & \makecell{Mixed-I} & Mixed-G & Mixed-S \\
    \hline
    \multirow{5}{*}{\rotatebox[origin=c]{90}{\makecell{\textbf{\cub}\\\textbf{(200 classes)}}}}
& Standard & 86.81 & 86.26 \decrease{0.55} & \textbf{87.04}\increase{0.23} &
87.02 & 89.18 & 88.89 \decrease{0.29} & 89.25 & \textbf{89.39}\increase{0.21} \\
    & No Background & 76.08 & \textbf{78.94}\increase{2.86} & 77.8 & 76.75 & 81.74 & \textbf{86.99}\increase{5.25} & 86.05 & 85.54 \\
    & No Birds & 5.51 & 3.45 & 5.29 & \textbf{5.99}\increase{0.48} & 6.02 & 5.99 & 4.85 & \textbf{6.68}\increase{0.66} \\
    & Small Boxes & \textbf{75.39} & 72.85 & 74.61 & 75.01 \decrease{0.38} & 86.9 & 87.57 & 87.06 & \textbf{87.78}\increase{0.88} \\
    & Big Box & 60.41 & 58.84 & 60.7 & \textbf{61.10}\increase{0.69} & 77.42 & 77.56 & 77.25 & \textbf{78.98}\increase{1.56} \\
    \hline
    \multirow{5}{*}{\rotatebox[origin=c]{90}{\makecell{\textbf{\nabirds}\\\textbf{(555 classes)}}}} & Standard & 80.23 & 79.21 \decrease{1.02} & \textbf{81.06}\increase{0.83} & 80.72 & 88.42 & 87.76 \decrease{0.66} & 88.67 & \textbf{88.75}\increase{0.33} \\
    & No Background & 65.18 & \textbf{69.18}\increase{4.00} & 67.14 & 67.36 & 80.83 & \textbf{85.99}\increase{5.16} & 84.83 & 84.32 \\
    & No Birds & 3.77 & 1.64 & 3.35 & \textbf{3.81}\increase{0.04} & 7.34 & 3.69 & 6.72 & \textbf{8.09}\increase{0.75} \\
    & Black Boxes & 61.5 & 59.77 & 61.67 & \textbf{61.82}\increase{0.32} & 85.21 & 84.48 & 85.37 & \textbf{85.47}\increase{0.26} \\
    & Big Box & 51.48 & 49.06 & \textbf{51.81}\increase{0.33} & 51.76 & 76.67 & 74.60 & 76.41 & \textbf{77.66}\increase{0.99} \\
    \hline
    \end{tabular}
    \label{tab:combined_performance}
\end{table*}

\paragraph{Dataset used in \unimodal models} We train and test models on splits of \cub and \nabirds, infusing habitat information into the training process using Mixed-S and Mixed-G methods. Test outcomes are recorded in \cref{tab:combined_performance}.

\begin{figure}[ht]
  \centering
    \scriptsize\makebox[\columnwidth][c]{Dryocopus Martius (Black Woodpecker)}%
    
    \begin{minipage}{0.9\columnwidth}
        \includegraphics[width=\columnwidth]{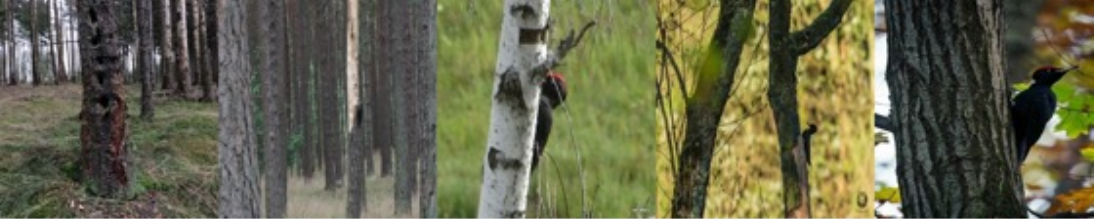}
    \end{minipage} 

    \scriptsize\makebox[\columnwidth][c]{Pelecanus Crispus (Dalmatian Pelican)}%
    
    \begin{minipage}{0.9\columnwidth}
        \includegraphics[width=\columnwidth]{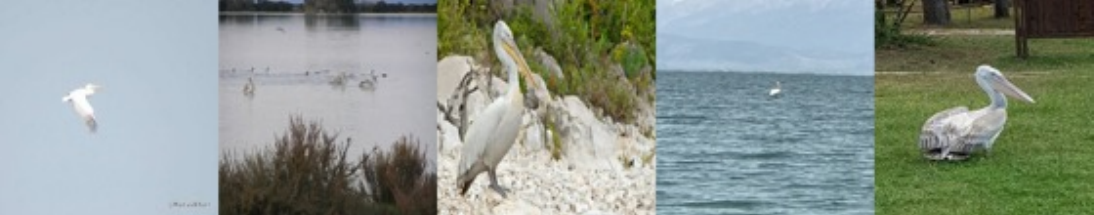}
    \end{minipage}
  
  \caption{Examples in \inat \cite{van2021benchmarking} where the birds \class{Dryocopus Martius (Black Woodpecker)} and \class{Pelecanus Crispus (Dalmatian Pelican)} being partially occluded by trees, poor lighting conditions, and the diminutive size of the bird.}
  \label{fig:inat_example}
\end{figure}

We also verify the advantage of habitats by testing trained models on the \inat dataset (1048 classes) featuring several challenges such as birds being partially occluded, or low lighting conditions (\cref{fig:inat_example}). Consequently, we derived two subsets, \inatcub (200 classes) and \inatnabirds (555 classes), to facilitate focused evaluations (see Tab. \ref{tab:multimodal_inat}), specifically:
\begin{itemize}
    \item \textbf{\inatcub} which includes 200 classes, with 144 overlapping classes both presenting in \inat and \cub and 56 classes sourced from \cub only.
    \item \textbf{\inatnabirds} which includes 555 classes, with 247 overlapping classes both presenting in \inat and \nabirds and 308 classes sourced from \nabirds only.
    
    Note that, images of overlapping classes are from \inat, while images of non-overlapping classes are sourced from either \cub or \nabirds.
\end{itemize}
The objective is to determine whether habitat information improves identification in real-world conditions where birds are occluded or appear diminutive. The results are detailed in \cref{tab:unimodal_inat_accuracy}.

\paragraph{Dataset used in CLIP}
\cub, \nabirds, and \inat are used to test the performance of CLIP with a keen focus on the role of habitat information. Habitat details for dataset classes are meticulously sourced from \textit{\allaboutbird}, a reputable avian resource. 

For \cub, which consists of 200 classes, comprehensive habitat information is directly obtained for 183 classes. For the remainder, we utilize GPT-3 to generate the missing habitat data.
 
In the case of \nabirds, the original 555 classes are streamlined to 267 classes for testing. This exclusion arises due to incomplete visual descriptions for classes that have variations like males, females, and juveniles.

Lastly, in \inat, out of 1486 classes, habitat information is successfully gathered for 425 classes, corresponding to the data available on \textit{\allaboutbird}. 

The impact of incorporating habitat descriptions on CLIP's zero-shot accuracy for \cub, \nabirds, and \inat is shown in \cref{tab:multimodal_cub_nabirds,tab:multimodal_inat}.
\subsection{Habitat Information consistently improves classification accuracy of \unimodal models}
To investigate the hypothesis that applying habitat-based augmentation (Mixed-S and Mixed-G) enhances the accuracy of bird identification, we trained unimodal models, which are ResNet-50 \cite{taesiri2022visual}, and TransFG \cite{he2022transfg}, with and without the augmentation strategies, and compare the models trained with augmentations against their non-augmented counterpart across different test sets of \cub and \nabirds as illustrated in \cref{fig:adversarial_examples}.

Furthermore, to investigate the robustness of our models under complex conditions - where birds may be partially obscured, or hidden by objects (see Fig. \ref{fig:inat_example}), we tested them on the challenging \inatcub and \inatnabirds datasets as described in \cref{sec:experiment}.

\paragraph{Experiment} First, we resize input images
to 256*256. We fine-tune ResNet-50 from \cite{taesiri2022visual} using a batch size of 64 and a learning rate of $1e^{-4}$
 over 20 epochs. For TransFG \cite{he2022transfg}, we employed a batch size of 8, maintained the learning rate at 
$1e^{-5}$, and extended training to 50 epochs. Adam \cite{kingma2014adam} optimizer is employed with a step learning rate scheduler. The pre-trained models are provided in \cref{sec:pretrained_model_links}.
In this experiment, three augmentation techniques Mixed-S, Mixed-G, and Mixed-I will be compared. For each type of augmentation, we test it with different test sets, as depicted in \cref{fig:adversarial_examples}, there are five types of test sets:
\begin{itemize}
    \item \textbf{Original}: Original test sets of \cub or \nabirds.
    \item \textbf{Black background}: Removing the background of the original set.
    \item \textbf{No Bird}: Crop the birds in the original set and use LaMa \cite{suvorov2022resolution} to fill in the cropped region.
    \item \textbf{Black Boxes}: Put eight small black boxes on the birds.
    \item \textbf{Big Box}: Put a big black box on the birds.
\end{itemize}

\paragraph{Results} The results in \cref{tab:combined_performance} show that \textbf{the models trained with Mixed-S and Mixed-G augmentations consistently enhance the accuracy of bird identification} which means habitat backgrounds really help in bird recognition. Concretely, using Mixed-G increases the accuracy by \increasenoparent{0.23} and \increasenoparent{0.83} points compared to not using any habitat augmentation methods when training ResNet-50 \cite{taesiri2022visual} on \cub, and \nabirds, respectively.
While training ViT \cite{he2022transfg} on those datasets, Mixed-S also gains \increasenoparent{0.21}, and \increasenoparent{0.33} points. Although the gaps are not substantial, they still suggest that classifiers trained on habitat-augmented datasets might be more effective at identifying birds. 

 The results in \cref{tab:unimodal_inat_accuracy} also suggest that models trained with Mixed-G and Mixed-S consistently achieve the highest accuracy on \inatcub and \inatnabirds. Concretely, in \inatcub, there is an average improvement of \increasenoparent{0.76} points in accuracy, while the improvement in \inatnabirds is \increasenoparent{0.48} points. This shows that in those cases, habitat background plays an important role in identifying a bird.

\begin{table}
\scriptsize
    \centering
    \caption{Performance comparison on \inatcub and \inatnabirds using VisualCorr \cite{taesiri2022visual} and TransFG \cite{he2022transfg} shows \AugGroup and \AugSame outperform \AugNone by \increasenoparent{0.76} and \increasenoparent{0.48} points, respectively, due to added 'habitat' in training.}
    \begin{tabular}{|c|c|c|c|c|}
    \hline
      & \multicolumn{2}{c|}{\makecell{\textbf{\inatcub}\\\textbf{(200 classes)}}} & \multicolumn{2}{c|}{\makecell{\textbf{\inatnabirds}\\\textbf{(555 classes)}}}  \\
    \hline
       & \textbf{ResNet-50} & \textbf{TransFG} & \textbf{ResNet-50} & \textbf{TransFG} \\
    \hline
    \AugNone & 68.23 & 71.38 & 68.59 & 77.93\\
    \hline
    \makecell{Mixed-I} & 68.17 & 71.45 & 66.95 & 76.39 \\
    \hline
    \makecell{Mixed-G} & 68.49 & \textbf{72.5} \increase{1.12} & \textbf{69.03} \increase{0.44} & 78.25 \\
    \hline
    \makecell{Mixed-S} & \textbf{68.62} \increase{0.39} & 71.89 & 68.63 & \textbf{78.45} \increase{0.52} \\
    \hline
    \textbf{Avg $\Delta$} & \multicolumn{2}{c|}{\increasenoparent{0.76}} & \multicolumn{2}{c|}{\increasenoparent{0.48}} \\ \hline
    \end{tabular}
    \label{tab:unimodal_inat_accuracy}
\end{table}

\subsection{Improving Zero-shot and Few-shot Accuracy with the help of Habitat Description}
\label{sec:clip_exp}
To assess habitat efficacy in CLIP, we compare the zero-shot and few-shot accuracies across \cub (200 classes), \nabirds (267 classes), and \inat (425 classes) both with and without integrating habitat descriptions into the visual concepts for each category.


\paragraph{Experiment}We augment each set of textual descriptions with habitat information, and examine the improvement of CLIP models. Additionally, in the few-shot context, our methodology involves the addition of 30 images for each class, specifically chosen at random from the training dataset.

Note that, we obtain the visual appearance descriptions from three sources, specifically: 
\begin{itemize}
    \item \textbf{M\&V} \cite{menon2022visual} utilizes GPT-3 \cite{brown2020language} to generate visual descriptions of birds. Sometimes it includes abstract features that are not presented in the image, such as \texttt{\footnotesize "This is a medium-sized bird"}, or \texttt{\footnotesize "wingspan of up to 3.6 m (12 ft)".} (Appendix \ref{sec:app_description}) 
    \item \textbf{PEEB} \cite{anonymous2023partbased} leverages GPT-4 \cite{openai2023gpt4} to generate descriptions of twelve distinct parts of birds including \texttt{\footnotesize wings, tail, eyes, back, forehead, nape, crown, leg, breast, throat, belly, beak}. (Appendix \ref{sec:app_description})
    \item \textbf{\ssc (SSC)} \cite{allaboutbirds} descriptions are human-annotated descriptions and are available on the \allaboutbird. (Appendix \ref{sec:app_description})
\end{itemize}

Furthermore, in order to be consistent between scientific class names in the \inat dataset (see \cref{fig:inat_example}), and the common bird names in the habitat description, we initially substitute common names in habitat descriptions with their corresponding scientific names. Subsequently, we inverted this process by replacing the scientific class names with common names. This approach was based on the premise that CLIP may not effectively recognize birds' scientific nomenclature, potentially impairing its accuracy in generating prompts that contain such terminology.

Motivated by \cite{menon2022visual}, we also modify the standard CLIP input prompt \texttt{\footnotesize "A photo of a \{c\}".} into \texttt{\footnotesize "\{c\}, which
(is/has/etc) \{description\}."} where \texttt{\footnotesize "\{c\}"} is a class name. Subsequently, the modified prompts can be fed into CLIP's text encoder.

\paragraph{Results}
We conduct comparisons of zero-shot and few-shot accuracy, considering the use of descriptions with and without habitat information. The results, as presented in \cref{tab:multimodal_cub_nabirds,tab:multimodal_inat,tab:few_shot_cub_nabirds}, clearly demonstrate that incorporating habitat descriptions leads to a notable improvement in both settings. For instance, in the zero-shot context, the combination of \ssc (SSC) and habitat achieves \increasenoparent{1.1} and \increasenoparent{0.99} pts improvement in \cub and \nabirds, respectively. Similarly, for the few-shot setting, using M\&V descriptions with habitat also gains \increasenoparent{4.63} and \increasenoparent{1.09} pts in these two datasets.




\begin{table}
\scriptsize
    \centering
    \caption{Zero-shot accuracy on \cub and \nabirds with M\&V, PEEB, and SSC descriptions in CLIP models shows habitat integration boosts accuracy across all descriptions. Average gains with habitat in SSC are \increasenoparent{1.1} for CUB and \increasenoparent{0.99} for \nabirds. Notably, adding habitat to M\&V descriptions in CLIP outperforms class names alone by \increasenoparent{2.73} points.}
    \begin{tabular}{|l|c|c|c|c|c|c|c|}
        \hline
        \multicolumn{8}{|c|}{\cellcolor{gray!25}\textbf{\cub (200 classes)}} \\ \hline
         & CLIP \cite{radford2021learning} & \multicolumn{2}{c|}{M\&V \cite{menon2022visual}} & \multicolumn{2}{c|}{PEEB \cite{anonymous2023partbased}} & \multicolumn{2}{c|}{SSC \cite{allaboutbirds}} \\ \cline{1-7} \cline{8-8}
        Habitat & \textcolor{red}\xmark & \textcolor{red}\xmark & \textcolor{green}\checkmark & \textcolor{red}\xmark & \textcolor{green}\checkmark & \textcolor{red}\xmark & \textcolor{green}\checkmark \\ \hline
        B/32 & 51.95 & 52.35 & \textbf{53.56} & 52.73 & \textbf{53.16} & 52.43 & \textbf{53.02} \\ \hline
        B/16 & 56.35 & 57.51 & \textbf{58.56} & 57.92 & \textbf{58.5} & 57.02 & \textbf{58.63} \\ \hline
        L/14 & 63.08 & 64.03 & \textbf{64.27} & 64.31 & \textbf{65.17} & 63.81 & \textbf{64.91} \\ \hline
        \multicolumn{2}{|c|}{\textbf{Avg}} & 57.96 & \textbf{58.80} & 58.32 & \textbf{58.94} & 57.75 & \textbf{58.85}  \\ \hline
        \multicolumn{2}{|c|}{\textbf{$\Delta$}} & \multicolumn{2}{c|}{\increasenoparent{0.83}} & \multicolumn{2}{c|}{\increasenoparent{0.62}} & \multicolumn{2}{c|}{\increasenoparent{1.1}}  \\ \hline
        \multicolumn{8}{|c|}{\cellcolor{gray!25}\textbf{\nabirds (267 classes that do not have annotation)}} \\ \hline
        B/32 & 49.08 & 49.79 & \textbf{51.17} & 50.61 & \textbf{51.25} & 49.84 & \textbf{50.84} \\ \hline
        B/16 & 55.28 & 57.29 & \textbf{58.06} & 56.87 & \textbf{57.59} & 56.12 & \textbf{57.07} \\ \hline
        L/14 & 63.86 & 66.56 & \textbf{67.17} & 66.73 & \textbf{67.33} & 64.98 & \textbf{66.01} \\ \hline
        
        \multicolumn{2}{|c|}{\textbf{Avg}} &57.88& \textbf{58.80} & 58.07 &	\textbf{58.72} & 56.98 & \textbf{57.97}  \\ \hline
        \multicolumn{2}{|c|}{\textbf{$\Delta$}} & \multicolumn{2}{c|}{\increasenoparent{0.92}} & \multicolumn{2}{c|}{\increasenoparent{0.65}} & \multicolumn{2}{c|}{\increasenoparent{0.99}}  \\ \hline
    \end{tabular}
    \label{tab:multimodal_cub_nabirds}
\end{table}

\begin{table}
\scriptsize
    \centering    
    \caption{Zeroshot Accuracy on \inat (425 classes) in two settings: common names to scientific names and vice versa. The results show habitat descriptions enhance zero-shot accuracy across all description sets, with common names notably yielding higher accuracy.}
    \begin{tabular}{|l|c|c|c|c|c|c|c|}
    \hline
        \multicolumn{8}{|c|}{\textbf{\cellcolor{gray!25}iNaturalist21 - only bird (425 classes, common names to scientific names)}} \\ \hline
         & CLIP & \multicolumn{2}{c|}{M\&V \cite{menon2022visual}} & \multicolumn{2}{c|}{PEEB \cite{anonymous2023partbased}} & \multicolumn{2}{c|}{SSC \cite{allaboutbirds}}\\ \cline{1-7} \cline{8-8}
        Habitat & \textcolor{red}\xmark & \textcolor{red}\xmark & \textcolor{green}\checkmark & \textcolor{red}\xmark & \textcolor{green}\checkmark & \textcolor{red}\xmark & \textcolor{green}\checkmark \\ \hline
        B/32 & 4.22 & 5.92 & \textbf{6.83} & 5.21 & \textbf{5.80} & 8.72 & \textbf{8.84} \\ \hline
        B/16 & 5.36 & 7.76 & \textbf{9.08} & 6.61 & \textbf{7.58} & 10.85 & \textbf{11.32} \\ \hline
        L/14 & 7.94 & 11.29 & \textbf{13.08} & 9.09 & \textbf{10.26} & 14.07 & \textbf{14.89} \\ \hline
        \multicolumn{2}{|c|}{\textbf{Avg}} &8.32&\textbf{9.66} & 6.97&\textbf{7.88}&11.21&\textbf{11.68}  \\ \hline
        \multicolumn{2}{|c|}{\textbf{$\Delta$}} & \multicolumn{2}{c|}{\increasenoparent{1.34}} & \multicolumn{2}{c|}{\increasenoparent{0.91}} & \multicolumn{2}{c|}{\increasenoparent{0.47}}  \\ \hline
        \multicolumn{8}{|c|}{\textbf{\cellcolor{gray!25}iNaturalist21 - only bird (425 classes, scientific names to common names)}} \\ \hline
        B/32 & 29.56 & 30.52 & \textbf{30.67} & 30.13 & \textbf{30.47} & 29.46 & \textbf{30.08} \\ \hline
        B/16 & 35.41 & 36.75 & \textbf{37.24} & 36.48 & \textbf{37.00} & 35.15 & \textbf{36.59} \\ \hline
        L/14 & 43.54 & 45.47 & \textbf{45.93} & 45.51 & \textbf{45.90} & 43.74 & \textbf{45.03} \\ \hline
        \multicolumn{2}{|c|}{\textbf{Avg}} & 37.58 &	\textbf{37.95} &	37.37 &	\textbf{37.79} &	36.12 &	\textbf{37.23}  \\ \hline
        \multicolumn{2}{|c|}{\textbf{$\Delta$}} & \multicolumn{2}{c|}{\increasenoparent{0.37}} & \multicolumn{2}{c|}{\increasenoparent{0.42}} & \multicolumn{2}{c|}{\increasenoparent{1.12}}  \\ \hline
    \end{tabular}
    \label{tab:multimodal_inat}
\end{table}

\begin{table}
\scriptsize
    \centering
    \caption{Few-shot accuracy on \cub and \nabirds with M\&V, PEEB, and SSC descriptions in CLIP models, reveals habitat integration enhances accuracy for all sets. Notably, habitat in M\&V descriptions improves by \increasenoparent{4.63} for \cub and \increasenoparent{1.09} for \nabirds.}
    \begin{tabular}{|l|l|l|l|l|}
    \hline 
        \multicolumn{5}{|c|}{\cellcolor{gray!25}\textbf{\cub (200 classes)}} \\ 
        \hline
        Description & B/32 & B/16 & L/14 & Avg $\Delta$\\ 
        \hline
        \multirow{2}{*}{\makecell{PEEB \cite{anonymous2023partbased}\\+Habitat}} & 59.39 & 71.51 & 78.13 & \multirow{2}{*}{\increasenoparent{1.25}}\\ \cline{2-4}
        & 62.93 \increase{3.54} & 71.66 \increase{0.15} & 78.18 \increase{0.05} & \\ \hline
        \multirow{2}{*}{\makecell{M\&V \cite{menon2022visual}\\+Habitat}} & 46.43 & 55.02 & 63.24 & \multirow{2}{*}{\increasenoparent{4.63}} \\ \cline{2-4}
        & 50.38 \increase{3.95} & 59.1 \increase{4.08} & 69.09 \increase{5.85} & \\ \hline
        \multirow{2}{*}{\makecell{SSC \cite{allaboutbirds}\\+Habitat}} & 58.75 & 69.16 & 76.94 & \multirow{2}{*}{\increasenoparent{0.56}} \\ \cline{2-4}
        & 59.37 \increase{0.62} & 69.76 \increase{0.60} & 77.39 \increase{0.45} & \\ \hline
        \multicolumn{5}{|c|}{\cellcolor{gray!25}\textbf{\nabirds (267 classes that do not have annotation)}} \\ \hline
        \multirow{2}{*}{\makecell{PEEB \cite{anonymous2023partbased}\\+Habitat}} & 57.29 & 67.65 & 76.06 & \multirow{2}{*}{\increasenoparent{0.19}} \\ \cline{2-4}
        & 57.55 \increase{0.26} & 67.77 \increase{0.12} & 76.25 \increase{0.19} & \\ \hline
        \multirow{2}{*}{\makecell{M\&V \cite{menon2022visual}\\+Habitat}} & 45.80 & 54.30 & 66.88 & \multirow{2}{*}{\increasenoparent{1.09}} \\ \cline{2-4}
        & 47.05 \increase{1.25} & 55.55 \increase{1.25} & 67.64 \increase{0.76} & \\ \hline
         \multirow{2}{*}{\makecell{SSC \cite{allaboutbirds}\\+Habitat}} & 53.57 & 65.65 & 74.73 & \multirow{2}{*}{\increasenoparent{0.44}} \\ \cline{2-4}
        & 54.13 \increase{0.56} & 66.17 \increase{0.52} & 74.96 \increase{0.23} & \\ \hline
    \end{tabular}
    \label{tab:few_shot_cub_nabirds}
\end{table}

\begin{figure}[ht]
  \centering
  \includegraphics[width=\columnwidth]{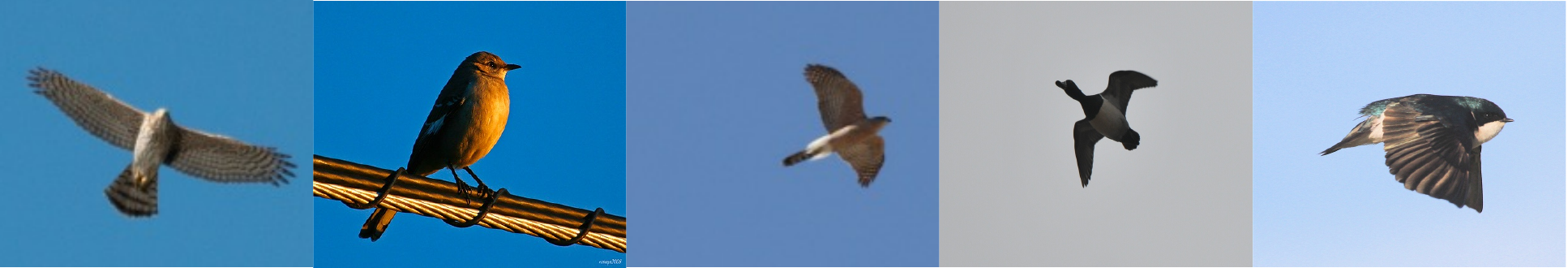}
  \caption{Flybird images do not have the habitat cues.}
  \label{fig:flybird}
\end{figure}

\subsection{Both \unimodal Models and \multimodal Models Encountering Similar Challenges in Bird Identification}

\label{sec:class_wise}

To further analyze the joint effects of habitat on bird identification for \unimodal and \multimodal, we study how incorporating habitat improves these models. That is, we wonder if both models will benefit from certain classes that strongly correlate with habitat.

\begin{figure*}
  \centering
  \includegraphics[width=2\columnwidth]{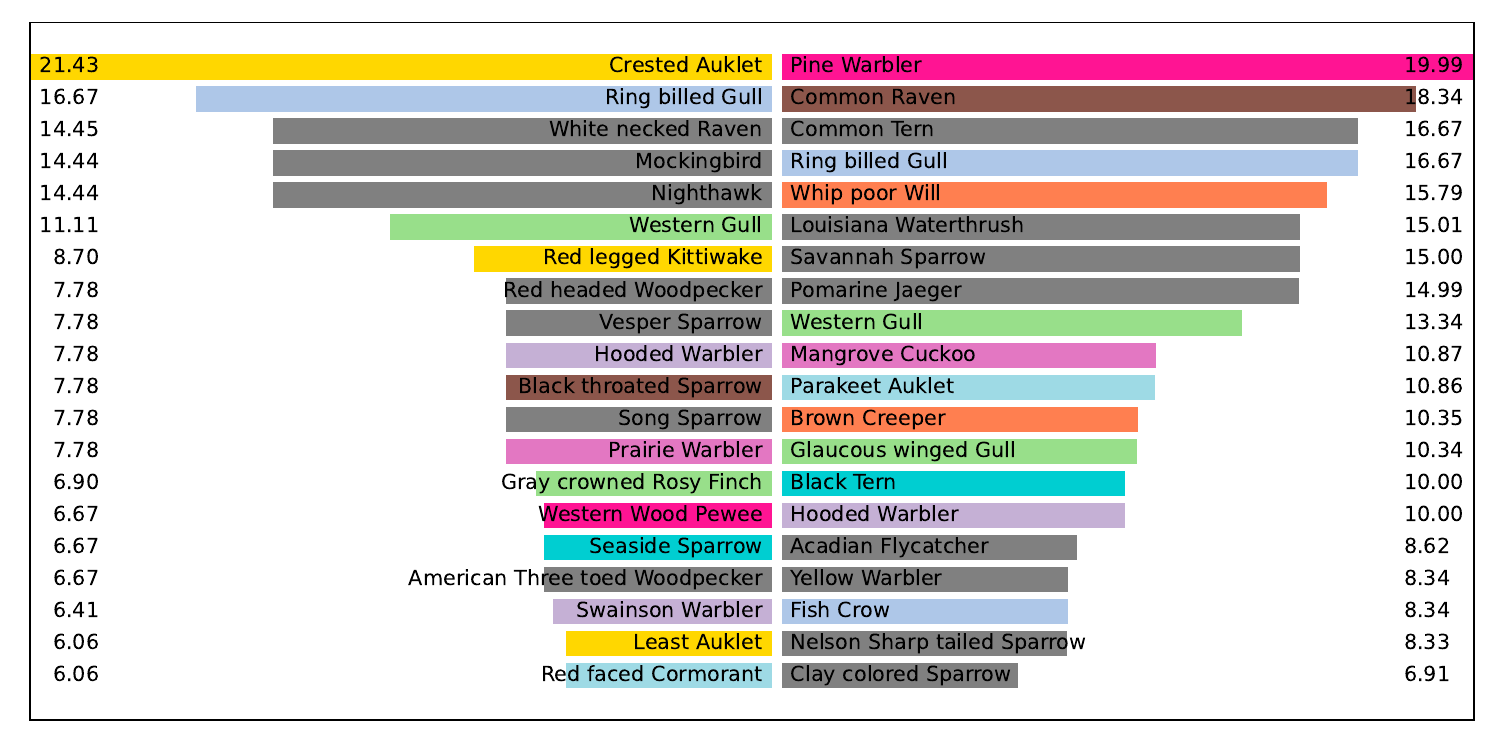}

  \makebox[1.\columnwidth][c]{Class-wise improvements using CLIP models}%
  \makebox[0.9\columnwidth][c]{Class-wise improvements using \unimodal models}%
  \caption{Accuracy improvement of the top 20 class-wise for CLIP and \unimodal models on \cub are shown in two charts, with colored bars indicating shared habitat groupings. Unique classes to each model without habitat matches are in \colorbox{gray}{gray}. Eight colors, excluding \colorbox{gray}{gray}, \colorbox{yellow}{yellow}, and \colorbox{orange}{orange}, highlight groups challenging for both models. CLIP struggles with \colorbox{yellow}{yellow-coded} birds, and \unimodal models with the \colorbox{orange}{orange-coded} group.}
  \label{fig:classwise_comparison}
\end{figure*}

\paragraph{Experiment}
We experiment on CUB by ranking the class-wise accuracy improvement of each class on \unimodal and \multimodal models separately (\cref{fig:classwise_comparison}). Then, based on the habitat group we proposed in \cref{sec:method_unimodal}, we analyze the classes that fall into the same habitat group in the top 20 classes for both models.

\paragraph{Results} 
As illustrated in \cref{fig:classwise_comparison}, among the top 20 classes, both models include over eight classes belonging to the habitat group, as indicated by the \colorbox{gray}{gray} color coding. This pattern suggests that both \unimodal and \multimodal models encounter similar challenges in identifying these habitat-related classes. Consequently, this observation emphasizes the significance of integrating habitat information into both \unimodal and \multimodal models for improved classification.

\subsection{Models perform poorly on Fly-bird images}
It is challenging to recognize the birds when they are flying \cite{wah2011caltech} (see Fig. \ref{fig:flybird}). With this premise, we design an experiment to investigate the performance of our models in such scenarios.

\paragraph{Fly Birds Detection} For identifying images that feature flying birds, we employ the Mask2Former model, as outlined in \cite{cheng2022masked}. This panoptic segmentation model first determines the presence of elements such as rocks, grass, or water in the image. Subsequently, it assesses whether the sky is depicted. If the sky is indeed present, the image is classified as containing flying birds; otherwise, it is categorized as a non-flying bird image.

\paragraph{Experiment} In this experiment, \cub and \nabirds are examined and we found that the fly-bird percentage of \cub is 10$\%$, while that of \nabirds is 11.4$\%$. Additionally, \cub and \nabirds are divided into \textbf{FlyBird-CUB, Non-FlyBird-CUB}, and \textbf{FlyBird-NABirds, non-FlyBird-NABirds}, and for all these datasets, we test the classification accuracy of our models.

\paragraph{Results} From \cref{tab:uncertain_1}, which showcases the results of \unimodal models, and \cref{tab:uncertain_2} which represents the performance of the CLIP model, several key observations can be made.

For both models, the Non-FlyBird results consistently achieve higher accuracy across almost all methods when compared to their FlyBird counterparts. Concretely, average improvements in \cub are \increasenoparent{1.31} and \increasenoparent{2.02} points, while in \nabirds the improvements are \increasenoparent{6.92} and \increasenoparent{4.08} points.

Further insight can be derived in that there is a big gap in accuracy between Non-FlyBird and FlyBird of \nabirds which are \increasenoparent{6.92}, and \increasenoparent{4.08} in \unimodal and CLIP, respectively. This is because there are 11.4\% FlyBirds images in this dataset, and the accuracy improves by a large margin when removing these images.

\begin{table}
\scriptsize
    \centering
    \caption{The table highlights the accuracy of various \unimodal models on \cub's FlyBird and Non-FlyBird (200 classes) and \nabirds (555 classes), revealing a consistent trend where Non-FlyBird outperforms FlyBird.}
    \begin{tabular}{|l|l|l|l|l|l|l|}
    \hline
        \multicolumn{2}{|c|}{\multirow{2}{*}{}} & \multicolumn{2}{c|}{\textbf{\cub (200 classes)}} & \multicolumn{2}{c|}{\textbf{\nabirds (555 classes)}} \\ \cline{3-6}
         \multicolumn{2}{|c|}{} & FlyBird & \makecell{Non-\\FlyBird}  & FlyBird & \makecell{Non-\\FlyBird}  \\ \hline
         
        \multirow{4}{*}[-0.ex]{\rotatebox[origin=c]{90}{ResNet-50}} & \AugNone & 86.06 & 86.93 \increase{0.87} & 72.68 & 81.16 \increase{8.48} \\ \cline{2-6}
        & \makecell{Mixed-S} & 85.89 & 87.21 \increase{1.32} & 72.96 & 81.68 \increase{8.72} \\ \cline{2-6}
        & \makecell{Mixed-G} & 83.97 & 86.55 \increase{2.58} & 73.96 & 81.95 \increase{7.99} \\ \cline{2-6}
        & \makecell{Mixed-I} & 85.71 & 87.27 \increase{1.56}  & 73.57 & 81.64 \increase{8.07} \\ \hline
        \multirow{4}{*}[-0.ex]{\rotatebox[origin=c]{90}{TransFG}} & \AugNone & 88.33 & 89.3 \increase{0.97} & 83.14 & 89.08 \increase{5.94} \\ \cline{2-6}
        & \makecell{Mixed-S} & 89.02 & 89.47 \increase{0.45} & 84.28 & 89.31 \increase{5.03} \\ \cline{2-6}
        & \makecell{Mixed-G} & 87.11 & 89.13 \increase{2.02} & 83.03 & 89.38 \increase{6.35} \\ \cline{2-6}
        & \makecell{Mixed-I} & 88.68 & 89.36 \increase{0.68} & 84.56 & 89.35 \increase{4.79} \\ \hline
        \multicolumn{2}{|c|}{Avg $\Delta$} & \multicolumn{2}{c|}{\increasenoparent{1.31}} & \multicolumn{2}{c|}{\increasenoparent{6.92}} \\ \hline
    \end{tabular}
    \label{tab:uncertain_1}
\end{table}
    
\begin{table}
\scriptsize
    \centering
    \caption{The table compares CLIP model performances using different descriptions on \cub's FlyBird and Non-FlyBird (200 classes) and \nabirds (267 classes). Descriptions combine habitat with SSC \cite{allaboutbirds}, M\&V \cite{menon2022visual}, or PEEB \cite{anonymous2023partbased}. A notable trend is Non-FlyBird's consistent outperformance over FlyBird.}

    \begin{tabular}{|l|l|l|l|l|l|}
    \hline
        \multicolumn{2}{|c|}{\multirow{2}{*}{}} & \multicolumn{2}{c|}{\textbf{\cub (200 classes)}} & \multicolumn{2}{c|}{\textbf{\nabirds (267 classes)}} \\ \cline{3-6}
        \multicolumn{2}{|c|}{} & FlyBird & Non-FlyBird & FlyBird & Non-FlyBird \\ \hline
        
        \multirow{3}{*}{\rotatebox[origin=c]{90}{ViT-B/32}} & \makecell{SSC\\+Habitat}& 48.43 & 53.36 \increase{4.93} & 45.75 & 51.5 \increase{5.75} \\ \cline{2-6}
        & \makecell{M\&V\\+Habitat}& 49.83 & 53.88 \increase{4.05} & 45.47 & 51.16 \increase{5.69} \\ \cline{2-6}
        & \makecell{PEEB\\+Habitat} & 49.83 & 53.42 \increase{3.59} & 46.46 & 51.63 \increase{5.17} \\ \hline
        \multirow{3}{*}{\rotatebox[origin=c]{90}{ViT-B/16}} & \makecell{SSC\\+Habitat} & 57.67 & 58.06 \increase{0.39} & 54.11 & 57.45 \increase{3.34} \\ \cline{2-6}
        & \makecell{M\&V\\+Habitat} & 58.01 & 58.61 \increase{0.6} & 54.18 & 58.09 \increase{3.91} \\ \cline{2-6}
        & \makecell{PEEB\\+Habitat} & 57.49 & 58.4 \increase{0.91} & 54.82 & 57.71 \increase{2.89} \\ \hline
        \multirow{3}{*}{\rotatebox[origin=c]{90}{ViT-L/14}} & \makecell{SSC\\+Habitat} & 64.29 & 64.52 \increase{0.23} & 62.96 & 66.41 \increase{3.45} \\ \cline{2-6}
        & \makecell{M\&V\\+Habitat} & 62.37 & 64.56 \increase{2.19} & 64.09 & 67.45 \increase{3.36} \\ \cline{2-6}
        & \makecell{PEEB\\+Habitat} & 63.59 & 64.90 \increase{1.31} & 64.90 & 67.43 \increase{3.19} \\ \hline
        \multicolumn{2}{|c|}{Avg $\Delta$} & \multicolumn{2}{c|}{\increasenoparent{2.02}} & \multicolumn{2}{c|}{\increasenoparent{4.08}} \\ \hline
    \end{tabular}
    \label{tab:uncertain_2}
\end{table}



\section{Conclusion and Future Work}
\label{sec:conclusion}

\paragraph{Limitation}
Our work suggests that including habitat information helps to improve bird recognition but getting correct habitat information for each bird is challenging due to the inherent variability in avian habitat, which may differ between nesting, foraging, and other activities, adding complexity to the task of obtaining precise and comprehensive habitat information.
Furthermore, it is essential to develop a novel methodology that quantifies the respective contributions of the habitat background and the bird's presence within an image, which may be beneficial for understanding cases where occlusion happens.

\paragraph{Future Work}
Part-based recognition systems hold promise as a future direction for improving bird recognition. By breaking down a bird's image into its constituent parts, such as beak, wings, tail, and body, habitat, and recognizing them individually, models can potentially enhance their accuracy and robustness.

Investigating the transferability of models trained with habitat-informed data to new geographic regions and diverse ecosystems also remains an essential aspect of future work.

\paragraph{Conclusion}
In conclusion, we have demonstrated the pivotal role of habitat information in enhancing the performance of both \unimodal models, including Convolutional Neural Networks (CNN) and Vision Transformers (ViT), as well as \multimodal models, with particular emphasis on CLIP in the zero-shot setting.



\clearpage
{\small

}

\clearpage
\setcounter{page}{1}

{
   \newpage
       \onecolumn
        \centering
        \Large
        \textbf{\paperTitle}\\
        \vspace{0.5em}Supplementary Material \\
        \vspace{1.0em}
   }


\newcommand{\beginsupplementary}{%
    \setcounter{table}{0}
    \renewcommand{\thetable}{A\arabic{table}}%
    
    \setcounter{figure}{0}
    \renewcommand{\thefigure}{A\arabic{figure}}%
    
    \setcounter{section}{0}
    \renewcommand{\thesection}{A\arabic{section}}
    \renewcommand{\thesubsection}{\thesection.\arabic{subsection}}
}
\beginsupplementary%




\section{Pretrained models}
\label{sec:pretrained_model_links}
\paragraph{Sources} We download the two pre-trained PyTorch models of ResNet-50, TransFG and three variants of CLIP (ViT/B-32, ViT/B-16, and ViT/L-14) from:
\begin{itemize}
    \item ResNet-50 \cite{taesiri2022visual}: \url{https://github.com/anguyen8/visual-correspondence-XAI/tree/main/ResNet-50}
    \item TransFG \cite{he2022transfg}: \url{https://github.com/TACJu/TransFG}
    \item CLIP \cite{radford2021learning}: \url{https://github.com/openai/CLIP}
\end{itemize}

\begin{figure*}[h]
  \centering
  \includegraphics[width=0.9\columnwidth]{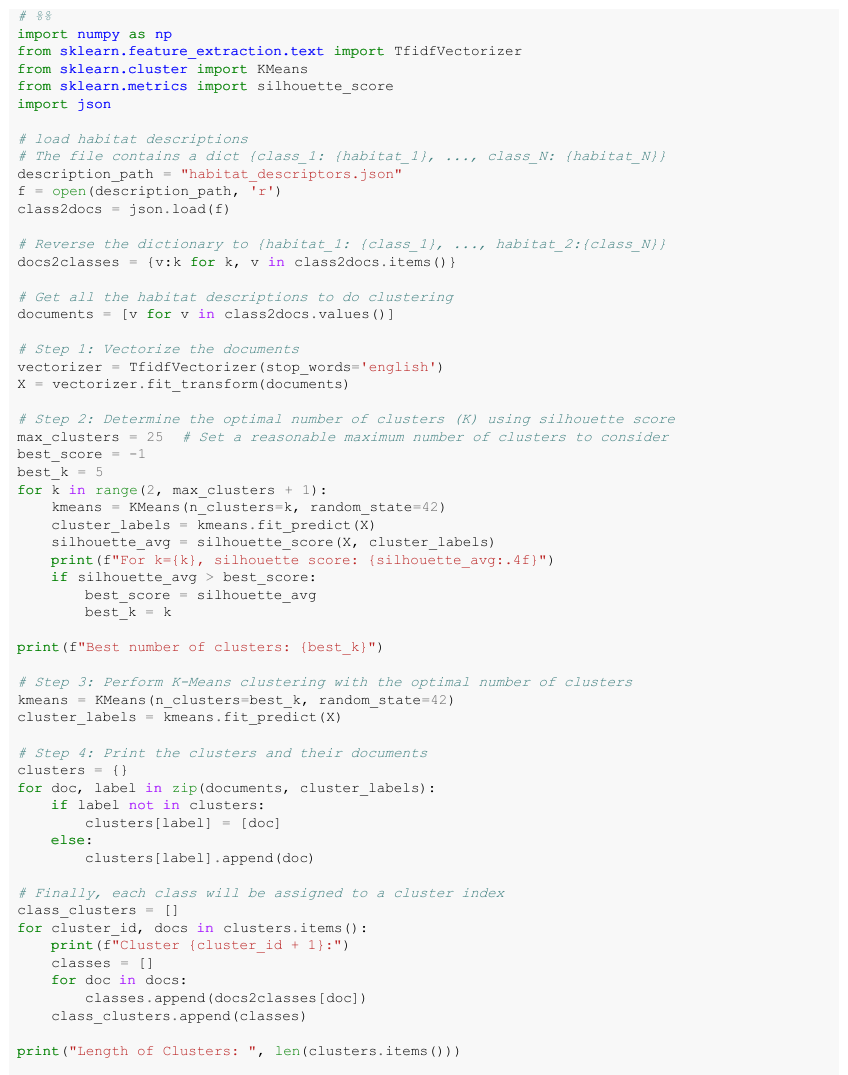}
  \caption{The algorithm used in this study to cluster bird species into groups based on their habitat settings.}
  \label{fig:text_clustering}
\end{figure*}

\section{Constructing Additional Data for Training \unimodal Models}
\label{sec:constructing_additional_data}
\subsection{Constructing Habitat Groups}
In our study, we implement a text clustering algorithm to analyze and group birds based on habitat descriptions (see \cref{fig:text_clustering}).

The algorithm begins by loading the textual descriptors, which are then transformed into a vector space using TF-IDF vectorization, excluding common English stop words for better feature representation. 

To determine the optimal number of clusters for k-means, we iteratively compute the silhouette scores for different cluster counts, selecting the cluster number that yields the highest score as the most suitable for our data. With the optimal number of clusters established, we perform k-means clustering and categorize the descriptions accordingly. Each cluster is then mapped back to the corresponding bird classes, resulting in a collection of class clusters.

As detailed in Section \ref{sec:method_unimodal}, identifying bird groups based on habitats facilitates both \AugGroup and \AugIrrelevant data augmentation processes. This categorization also enables a more nuanced analysis of class-wise accuracy between CLIP and unimodal models, which is further explored in Section \ref{sec:class_wise}.

For the \cub dataset, we organize 200 classes into 50 groups, whereas for the \nabirds dataset, 555 classes were clustered into 196 groups.

\subsection{Example of Bird Groups and their corresponding habitat descriptions}
\label{sec:example_bird_groups}
\begin{figure*}
  \centering  
  \makebox[\columnwidth][c]{\textbf{Group 1}: "Black-footed Albatross", "Heermann Gull", "Elegant Tern"}%
  
  \makebox[\columnwidth][c]{\textbf{Prefered Habitat}: far offshore, sandy island.}%
  
    \begin{minipage}{0.375\columnwidth}
        \includegraphics[width=\columnwidth]{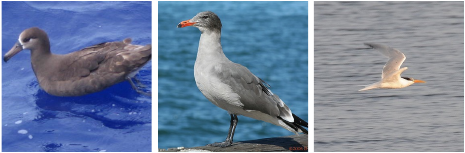}
    \end{minipage} 

    \makebox[\columnwidth][c]{\textbf{Group 2}: "Black-throated Sparrow", "Common Raven", "Green-tailed Towhee"}%

    \makebox[\columnwidth][c]{\textbf{Prefered Habitat}: scrubby areas, desert scrub.}%
    
    \begin{minipage}{0.375\columnwidth}
        \includegraphics[width=\columnwidth]{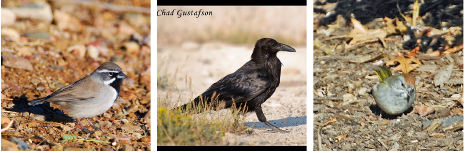}
    \end{minipage} 

    \makebox[\columnwidth][c]{\textbf{Group 3}: "Mallard", "Seaside Sparrow", "Barn Swallow", "Black Tern"}%

    \makebox[\columnwidth][c]{\textbf{Prefered Habitat}: ponds, fresh and salt water marshes}%
    
    \begin{minipage}{0.5\columnwidth}
        \includegraphics[width=\columnwidth]{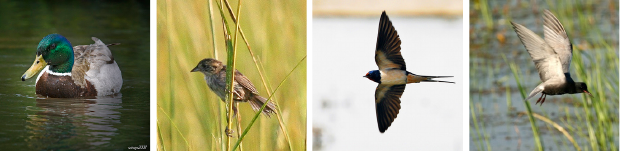}
    \end{minipage}

    \makebox[\columnwidth][c]{\textbf{Group 4}: "White-necked Raven", "Geococcyx", "Sage Thrasher", "Rock Wren"}%

    \makebox[\columnwidth][c]{\textbf{Prefered Habitat}: Desert regions.}%
    
    \begin{minipage}{0.5\columnwidth}
        \includegraphics[width=\columnwidth]{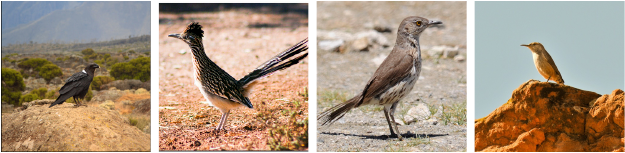}
    \end{minipage}

    \makebox[\columnwidth][c]{\textbf{Group 5}: "American Crow", "Fish Crow", "American Goldfinch", "Ring-billed Gull"}%

    \makebox[\columnwidth][c]{\textbf{Prefered Habitat}: agricultural fields, roadsides.}%
    
    \begin{minipage}{0.5\columnwidth}
        \includegraphics[width=\columnwidth]{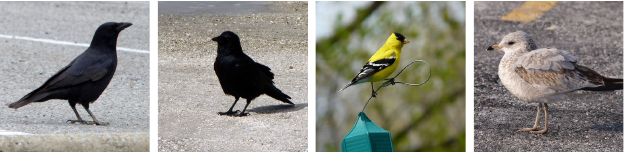}
    \end{minipage}

    \makebox[\columnwidth][c]{\textbf{Group 6}: "Brown Creeper", "Whip-poor Will", "Scarlet Tanager", "Red-eyed Vireo", "Black-and-white Warbler", "Black-throated-Blue Warbler"}%

    \makebox[\columnwidth][c]{\textbf{Prefered Habitat}: deciduous forests.}%
    
    \begin{minipage}{0.75\columnwidth}
        \includegraphics[width=\columnwidth]{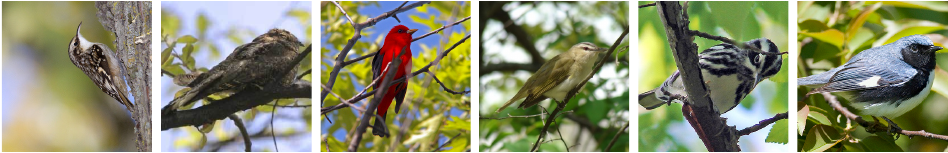}
    \end{minipage}

    \makebox[\columnwidth][c]{\textbf{Group 7}: "Gray-crowned-Rosy Finch", "Pigeon Guillemot", "Glaucous-winged Gull", "Ivory Gull", "Slaty-backed Gull", "Western Gull"}%

    \makebox[\columnwidth][c]{\textbf{Prefered Habitat}: cliffs, ice field, maritime location.}%
    
    \begin{minipage}{0.75\columnwidth}
        \includegraphics[width=\columnwidth]{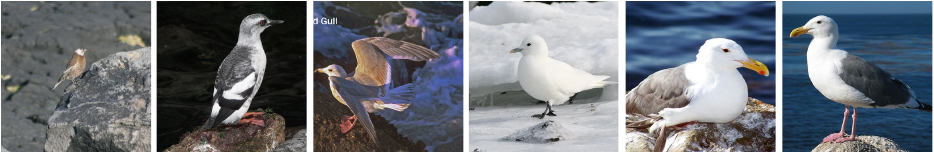}
    \end{minipage}
    
  \caption{Seven Bird Species Groups - Each group consists of birds sharing a common habitat. For example, the 7th group includes \class{Gray-crowned-Rosy Finch, Pigeon Guillemot, Glaucous-winged Gull, Ivory Gull, Slaty-backed Gull, and Western Gull}, they all prefer cliffs, mountains or beaches.}
  \label{fig:bird_groups}
\end{figure*}

The group containing the \class{Black-footed Albatross, Heerman Gull, and Elegant Tern}, usually prefers far offshore or sandy island (see \cref{fig:bird_groups} at row 1).
\begin{itemize}
    \item \textbf{Black-footed Albatross}: Nests on low, sandy islands in the tropical North Pacific. Forages both near shore (though usually not within sight of land) and far offshore, in places where upwelling or converging currents concentrate nutrients and prey at the sea surface. \\
    \item \textbf{Heerman Gull}: Nests on rocky islands, mostly in the Gulf of California. Forages in ocean waters, usually within sight of land, often with terns, pelicans, cormorants, boobies, and sea lions. Also forages along beaches and in sheltered harbors and estuaries. Very rare inland.\\
    \item \textbf{Elegant Tern}: Nests on beaches and sandy islands. Forages close to the shore over ocean waters, where currents and upwelling concentrate prey (northern anchovy in particular). Usually forages within 10 miles of land, and often within sight of land.
\end{itemize}
Group containing the \class{Mallard, Seaside Sparrow, Barn Swallow, and Black Tern}, their hangouts usually are ponds, fresh and saltwater marshes (see \cref{fig:bird_groups} at row 3).
\begin{itemize}
    \item \textbf{Mallard}: Mallards can live in almost any wetland habitat, natural or artificial. Look for them on lakes, ponds, marshes, rivers, and coastal habitats, as well as city and suburban parks and residential backyards.\\
    \item \textbf{Seaside Sparrow}: Salt marshes, including brackish marshes and (in the Everglades) freshwater marshes.\\
    \item \textbf{Barn Swallow}: You can find the adaptable Barn Swallow feeding in open habitats from fields, parks, and roadway edges to marshes, meadows, ponds, and coastal waters. Their nests are often easy to spot under the eaves or inside of sheds, barns, bridges and other structures.\\
    \item \textbf{Black Tern}: Nests in freshwater marshes and bogs; winters in coastal lagoons, marshes, and open ocean waters. Migrants may stop over in almost any type of wetland.\\
\end{itemize}

\subsection{Augmenting additional data}
The core aspect of our proposed augmentation technique involves cropping birds from their original images and superimposing them onto various backgrounds, a process detailed in \cref{al:overlay_alg}.

\begin{algorithm}
\scriptsize
\caption{Pseudo code for combining a habitat image with a specific bird.}
\label{al:overlay_alg}
\begin{algorithmic}[1]
 \STATE \textbf{Input:} $habitat\_img$, $only\_bird\_img$, $mask$.
    \STATE \quad $habitat\_img$: an image after cropping the bird and inpainting the missing regions using LaMa \cite{suvorov2022resolution}.
    \STATE \quad $only\_bird\_img$: an image of a bird with a black background (bird only).
    \STATE \quad $mask$: binary mask of the bird in the $only\_bird\_img$.
    \STATE \quad $habitat\_masked$: binary mask of habitat image.
    \STATE \quad $combined\_img$: Combined image between the $habitat\_img$ and the bird in $only\_bird\_img$.
    \STATE \textbf{Output:} $combined\_img$
    \STATE $habitat\_masked\ \gets$ round($habitat\_img\ \times (1 - mask)$)
    \STATE $img\_combined\ \gets$ $black\_img$ + $habitat\_masked$
    \STATE \textbf{Return:} $combined\_img$
\end{algorithmic}
\end{algorithm}

\textbf{Mixed-Same} We implement a random alteration of the bird's background while ensuring that it remains within the same class as the original.

\textbf{Mixed-Group} Under this scenario, the background for a bird is carefully chosen from its own bird group. This selection process is elaborated in Section \ref{sec:example_bird_groups}.

\textbf{Mixed-Irrelevant} In the Mixed-Irrelevant case, the bird's background is randomly selected from groups that are irrelevant or unrelated to the bird’s original group.

Having established the sources of the habitat images, we can readily apply Algorithm \ref{al:overlay_alg} to generate additional data.

\section{Descriptions Types in CLIP}
\label{sec:app_description}
\paragraph{Visual descriptions}The visual descriptions used in this study were obtained from three sources:
\begin{itemize}
    \item \textbf{M\&V} \cite{menon2022visual} utilizes GPT-3 \cite{brown2020language} to generate visual descriptions of birds. For instance, the class \class{Ovenbird} has descriptions:
    
    \begin{lstlisting}[style=nohighlight]
    Ovenbird: {
    It is a small, sparrow-like bird,
    It is brown or grey with streaks on its breast,
    It has a white belly,
    It has a black stripe on its head,
    It has a long, curved beak,
    It has dark eyes,
    It has long legs
    }
    \end{lstlisting}

    \item \textbf{PEEB} \cite{anonymous2023partbased} leverages GPT-4 \cite{openai2023gpt4} to generate descriptions of twelve distinct parts of birds including \texttt{\footnotesize wings, tail, eyes, back, forehead, nape, crown, leg, breast, throat, belly, beak}. For instance, the class \class{Orange-crowned Warbler} has descriptions:

    \begin{lstlisting}[style=nohighlight]
    Orange-crowned Warbler: {
        back: olivegreen with darker streaks,
        beak: small and pointed,
        belly: pale yellowishgreen,
        breast: pale yellowishgreen with faint streaks,
        crown: dull olivegreen,
        forehead: dull olivegreen,
        eyes: dark brown,
        legs: dark grayishbrown,
        wings: olivegreen with two pale wing bars,
        nape: olivegreen with faint streaks,
        tail: brownishgray with white outer feathers,
        throat: pale yellowishgreen with faint streaks
    }
    \end{lstlisting}
    
    \item \textbf{\ssc (SSC)} \cite{allaboutbirds} descriptions are human-annotated descriptions and are available on the \allaboutbird. For instance, the class \class{Chestnut-sided Warbler} has descriptions:
    \begin{lstlisting}[style=nohighlight]
    Chestnut-sided Warbler: {
    shape: An slim warbler with a relatively long tail that it often holds cocked,
    or raised above the body line, which makes the tail appear longer still,
    size: Larger than a Ruby-crowned Kinglet, smaller than a Song Sparrow 
    sparrow-sized or smaller,
    color: Breeding adults are crisp gray-and-white birds with a yellow crown,
    black face markings, and rich chestnut flanks. Males are more richly marked than 
    females. In nonbreeding plumage, adults and immatures are bright lime-green 
    above with a neat white eyering, two wingbars, and pale gray to white underparts
    }
    \end{lstlisting}
    
\end{itemize}
\paragraph{Habitat descriptions}Additionally, habitat descriptions were also sourced from \allaboutbird \cite{allaboutbirds}. For instance: \\
The class \class{Red-headed Woodpecker} has habitat description: \\
\begin{lstlisting}[style=nohighlight]
Red-headed Woodpeckers: {
Red-headed Woodpeckers live in pine savannahs and other open forests with clear 
understories. Open pine plantations, treerows in agricultural areas, and standing 
timber in beaver swamps and other wetlands all attract Red-headed Woodpeckers
}
\end{lstlisting}
Or, The class \class{Cactus Wren} has habitat description: \\
\begin{lstlisting}[style=nohighlight]
Cactus Wren: {
Cactus Wrens live in deserts, arid foothills, coastal sage scrub, and urban areas
throughout the Southwestern deserts, especially in areas with thorny shrubs, cholla, 
and prickly pear
}
\end{lstlisting}

\subsection{Differentiating Habitats of Visually Similar Birds}
\label{sec:more_bird_pair}
\begin{figure}[ht]
  \centering
  \begin{subfigure}[t]{0.33\columnwidth}
    \includegraphics[width=\linewidth]{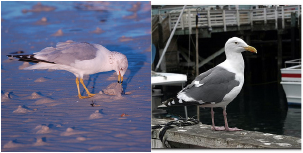}
    \scriptsize
    \begin{tabular}{c@{\hspace{0.8cm}}c}
    \hspace{0.4cm}\textbf{Ring-billed Gull} & \textbf{Western Gull} \\
    Beaches & Maritime location \\
    \end{tabular}
  \end{subfigure}
  \hfill 
  \begin{subfigure}[t]{0.33\columnwidth}
    \includegraphics[width=\linewidth]{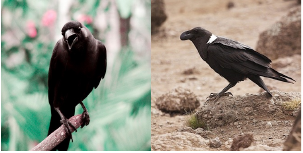}
    \scriptsize
    \begin{tabular}{c@{\hspace{0.8cm}}c}
    \hspace{0.4cm}\textbf{Common Raven} & \textbf{White-necked Raven} \\
    Forest & Desert Regions \\
    \end{tabular}
  \end{subfigure}
  \begin{subfigure}[t]{0.33\columnwidth}
    \includegraphics[width=\linewidth]{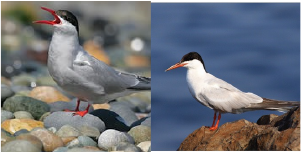}
    \scriptsize
    \begin{tabular}{c@{\hspace{1.6cm}}c}
    \hspace{0.6cm}\textbf{Artic Tern} & \textbf{Common Tern} \\
       \hspace{0.6cm}Treeless Area & Rocky Island \\
    \end{tabular}
  \end{subfigure}
  
  \begin{subfigure}[t]{0.33\columnwidth}
    \includegraphics[width=\linewidth]{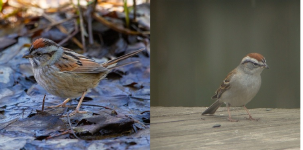}
    \scriptsize
      \begin{tabular}{c@{\hspace{0.8cm}}c}
    \hspace{0.5cm}\textbf{Swamp Sparrow} & \textbf{Chipping Sparrow} \\
    \hspace{0.8cm}Swamp & Common Backyard \\
    \end{tabular}
  \end{subfigure}
  \hfill 
  \begin{subfigure}[t]{0.33\columnwidth}
    \includegraphics[width=\linewidth]{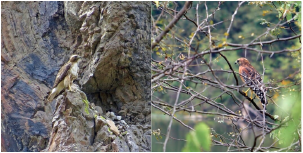}
    \scriptsize
      \begin{tabular}{c@{\hspace{0.8cm}}c}
    \hspace{0.4cm}\textbf{Red-tailed Hawk} & \textbf{Red-shoulder Hawk} \\
    \hspace{0.8cm}Mountain & Near River \\
    \end{tabular}
  \end{subfigure}
  \begin{subfigure}[t]{0.33\columnwidth}
    \includegraphics[width=\linewidth]{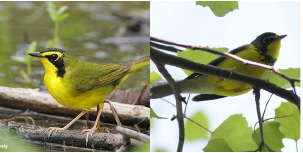}
    \scriptsize
      \begin{tabular}{c@{\hspace{0.6cm}}c}
    \hspace{0.4cm}\textbf{Kentucky Warbler} & \textbf{Canada Warbler} \\
    \hspace{0.6cm}Near Streams & Deciduous Forests \\
    \end{tabular}
  \end{subfigure}

  \caption{Comparative visual analysis of two bird species pairs, each exhibiting similar morphology yet distinct habitats. For example, the \class{Swamp Sparrow} (a) resides in swamps, contrasted with the \class{Chipping Sparrow} (b) which favors the common backyard. On the right, the \class{Red-tailed Hawk} (c) is adapted to mountains, whereas the \class{Red-shoulder Hawk} (d) is typically found near rivers.}
  \label{fig:more_bird_pair}
\end{figure}

\cref{fig:more_bird_pair} shows 6 pairs of birds where birds in each pair are visually similar but they have different habitat hangouts.

\end{document}